\documentclass[sigconf]{acmart}

\AtBeginDocument{%
  }

\setcopyright{acmlicensed}
\copyrightyear{2026}
\acmYear{2026}
\acmDOI{XXXXXXX.XXXXXXX}

\acmConference[MM'26]{Make sure to enter the correct
  conference title from your rights confirmation email}{November 10–14,
  2026}{Rio de Janeiro, Brazil}

\renewcommand\footnotetextcopyrightpermission[1]{}
\acmISBN{XXX-X-XXXX-XXXX-X/2026/10}
\usepackage{booktabs}   
\usepackage{tabularx}   
\usepackage{amsmath}    
\usepackage{array}       
\usepackage{ragged2e}    
\usepackage{graphicx} 
\usepackage{makecell}
\usepackage{multirow}
\usepackage{float} 
\usepackage{caption}
\usepackage{subcaption} 

\begin{document}

\title{BiTDiff: Fine-Grained 3D Conducting Motion Generation \\ via BiMamba-Transformer Diffusion}

\author{Tianzhi Jia}
\authornote{Equal Contribution.}
\email{jiatianzhi@bjtu.edu.cn}
\affiliation{%
    \institution{
    Institute of Information Science,
    Beijing Jiaotong University
    }
    \city{Beijing}
    \country{China}
}

\author{Kaixing Yang}
\authornotemark[1]
\email{yangkaixing@ruc.edu.cn}
\affiliation{%
    \institution{
        Renmin University of China
    }
    \city{Beijing}
    \country{China}    
}

\author{Xiaole Yang}
\authornotemark[1]
\email{yangxiaole6767@gmail.com}
\affiliation{%
    \institution{
        ADVANCE.AI
    }
    \city{Beijing}
    \country{China}
}

\author{Xulong Tang}
\email{Xulong.Tang@maloutech.com}
\affiliation{%
    \institution{
        Malou Tech Inc
    }
    \city{Plano, Texas}
    \country{USA}
}

\author{Ke Qiu}
\email{ke.qiu@maloutech.com}
\affiliation{%
    \institution{
        Malou Tech Inc
    }
    \city{Plano, Texas}
    \country{USA}
}

\author{Shikui Wei}
\authornote{Corresponding author.}
\email{shkwei@bjtu.edu.cn}

\author{Yao Zhao}
\email{yzhao@bjtu.edu.cn}

\affiliation{%
  \institution{
  Institute of Information Science, 
  Beijing Jiaotong University}
  \city{Beijing}
  \country{China}
}

\renewcommand{\shortauthors}{Tianzhi Jia et al.}

\begin{abstract}
3D conducting motion generation aims to synthesize fine-grained conductor motions from music, with broad potential in music education, virtual performance, digital human animation, and human-AI co-creation. However, this task remains underexplored due to two major challenges: (1) the lack of large-scale fine-grained 3D conducting datasets and (2) the absence of effective methods that can jointly support long-sequence generation with high quality and efficiency. To address the data limitation, we develop a quality-oriented 3D conducting motion collection pipeline and construct \textbf{CM-Data}, a fine-grained SMPL-X dataset with about 10 hours of conducting motion data. To the best of our knowledge, \textbf{CM-Data} is the first and largest public dataset for 3D conducting motion generation. To address the methodological limitation, we propose \textbf{BiTDiff}, a novel framework for 3D conducting motion generation, built upon a BiMamba-Transformer hybrid model architecture for efficient long-sequence modeling and a Diffusion-based generative strategy with human-kinematic decomposition for high-quality motion synthesis. Specifically, \textbf{BiTDiff} introduces auxiliary physical-consistency losses and a hand-/body-specific forward-kinematics design for better fine-grained motion modeling, while leveraging BiMamba for memory-efficient long-sequence temporal modeling and Transformer for cross-modal semantic alignment. In addition, \textbf{BiTDiff} supports training-free joint-level motion editing, enabling downstream human-AI interaction design. Extensive quantitative and qualitative experiments demonstrate that \textbf{BiTDiff} achieves state-of-the-art (SOTA) performance for 3D conducting motion generation on the \textbf{CM-Data} dataset. Code will be available upon acceptance.
\end{abstract}



\begin{teaserfigure}
  \centering
  \includegraphics[width=\textwidth]{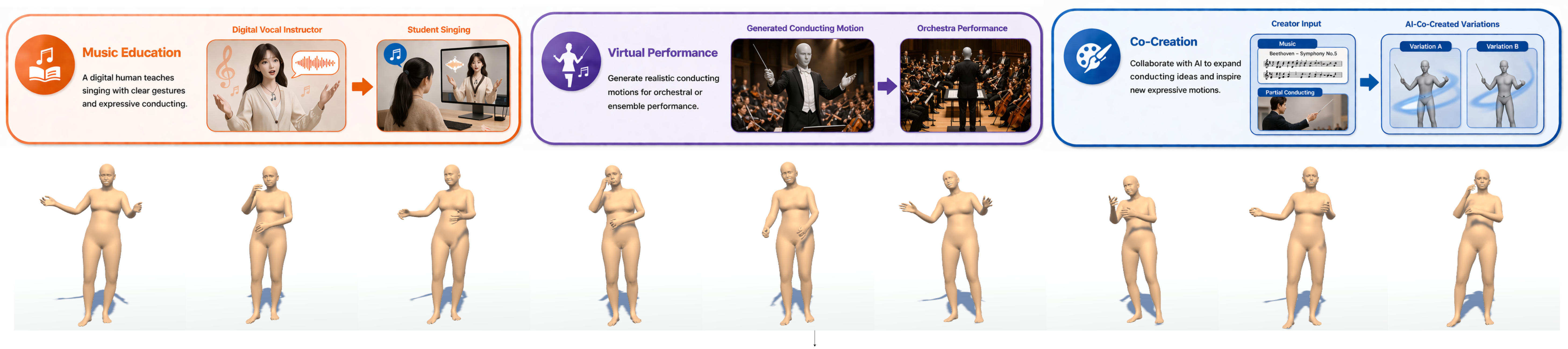}
  \vspace{-0.35in}
    \caption{\textbf{3D conducting motion generation} is a promising research direction with broad applications (up), such as \textbf{Music Education}, \textbf{Virtual Performance}, and \textbf{Co-Creation}. Moreover, the conducting motions generated by \textbf{BiTDiff} (low), trained on our proposed dataset \textbf{CM-Data}, are not only temporally coherent and rhythmically aligned with music, but also finely detailed and artistically expressive.}
  \label{fig:teaser}
  \vspace{0.2in}
\end{teaserfigure}

\maketitle
\section{Introduction}

Conducting motion serves as a crucial visual language in musical performance, enabling conductors to communicate tempo, dynamics, phrasing, and expressive intent to performers through body movements. Beyond its fundamental role in orchestra rehearsal and live performance, conducting motion also holds broad application value in areas such as music education, virtual performance, digital human animation, and human–AI co-creation~\cite{zhao2023taming,liu2022self}, as shown in Fig. ~\ref{fig:teaser}. With the rapid progress of 3D human motion recovery~\cite{wang2025prompthmr,zhang2025robust} and AI-generated content (AIGC)~\cite{yang2025mace,siyao2022bailando,yang2024beatdance}, data-driven analysis and synthesis of fine-grained conducting gestures have become increasingly feasible, making 3D conducting motion generation an emerging topic with broad potential in artistic expression and intelligent multimedia applications.

In recent years, substantial progress has been made in several research directions related to conducting motion generation, including 3D gesture generation~\cite{peng2023selftalk,zhang2025robust,zhang2025semtalk}, 3D dance generation~\cite{tseng2023edge,siyao2022bailando,yang2025megadance,yang2025flowerdance} and conducting motion generation~\cite{zhao2023taming,liu2022self,oh2024transfer}. Howerver, speech-driven 3D gesture generation primarily aim to extract speech-related semantic, while overlooking the music-structured control signals required in conducting motion generation, such as beat organization, ictus timing, and cueing. Similarly, music-driven 3D dance generation mainly focus on body-level music–motion alignment, while overlooking the fine-grained hand, upper-body, and facial control signals required in conducting motion generation.

A few studies have also explored conducting motion generation. ~\cite{liu2022self} first introduced a large-scale open-source dataset based on 2D keypoints. However, this dataset is relatively coarse, as it does not capture fine-grained head and hand details and cannot be readily generalized to 3D settings. ~\cite{zhao2023taming} proposed a diffusion-based model, while ~\cite{oh2024transfer} attempted to transfer the capability of 3D dance models to the conducting motion generation scenario; however, their generation quality and efficiency still fall short of industrial requirements, not to mention in the more challenging setting of long-sequence generation.
\textit{Overall, the field of 3D conducting motion generation currently faces two major challenges: (1) the lack of a large-scale fine-grained open-source dataset covering diverse conducting scenarios; and (2) the lack of effective methodology that can support long-sequence generation with high quality and efficiency.}

To tackle the dataset limitation, we develop a quality-oriented 3D conducting motion collection pipeline and build a large-scale 3D conducting motion dataset, termed \textbf{CM-Data}. Specifically, we design a deep-learning-based recording and processing workflow for high-fidelity 3D conducting motion capture. On the data side, we manually curate about 15 hours of videos that are more amenable to model learning, favoring stable viewpoints, high visibility, limited shot changes, clean lighting, and clear conductor prominence, thereby reducing uncontrolled noise caused by occlusion and domain shift at the source. On the reconstruction side, we decompose high-quality SMPL-X~\cite{pavlakos2019expressive} recovery into several specialized subproblems and address them with dedicated models: PromptHMR~\cite{wang2025prompthmr} for body reconstruction, HaPTIC~\cite{ye2025predicting} for detailed hand recovery, and SPECTRE~\cite{filntisis2022visual} for facial expression and deformation modeling. These components are then unified into a single fusion pipeline to produce high-fidelity SMPL-X motion sequences. 
In total, we obtain about 10 hours of fine-grained 3D SMPL-X data. To the best of our knowledge, \textbf{CM-Data} is the first and largest public datasets for 3D conducting motion generation. It covers both orchestral and choral conducting scenarios, with broad diversity in musical genre, ensemble type, and performance setting, and provides detailed hand, face, and full-body motion annotations, offering a stronger data foundation for fine-grained and long-horizon 3D conducting motion generation.

To tackle the methodological limitation, we propose \textbf{BiTDiff}, a novel framework for 3D conducting motion generation, built upon a BiMamba-Transformer hybrid model architecture for efficient long-sequence modeling and a Diffusion-based generative strategy with human-kinematic decomposition for high-quality motion synthesis.
For the generative strategy, we adopt diffusion as the core paradigm and further introduce auxiliary losses, following~\cite{tseng2023edge}, to enhance physical consistency during training. Moreover, to avoid underconstraining hand motions in a naive FK loss, we decompose the FK constraint into hand-specific and body-specific terms, improving fine-grained hand modeling while preserving overall body coherence. Furthermore, we introduce a training-free motion editing strategy during sampling, enabling joint/temporal-level motion manipulation without additional training and thereby effectively supporting downstream human-AI interaction design.
For the model architecture, we combine BiMamba and Transformer to leverage their complementary strengths: the Transformer is used to capture cross-modal global semantic information, while BiMamba is responsible for modeling efficient intra-modal temporal dynamics. Unlike autoregressive generation, this architecture supports a non-autoregressive generation process, which mitigates long-horizon drift caused by exposure bias~\cite{tseng2023edge,yang2025megadance,siyao2022bailando}. Benefiting from the linear-time complexity and scalability of Mamba~\cite{gu2023mamba}, the proposed architecture is also memory-efficient, making it particularly suitable for long-sequence generation. In addition, the bidirectional design of BiMamba alleviates the limitation of standard Mamba in modeling only one-directional context, while introducing only modest computational overhead.

In conclusion, our contributions are as follows:
\begin{itemize}
    \item We introduce a quality-oriented 3D conducting motion collection pipeline and construct \textbf{CM-Data}, a fine-grained 3D SMPL-X dataset with about 10 hours of conducting motion data. To the best of our knowledge, \textbf{CM-Data} is the first and largest public dataset for 3D conducting motion.
    
    \item We propose \textbf{BiTDiff}, a novel framework for 3D conducting motion generation, built upon a BiMamba-Transformer hybrid model architecture for efficient long-sequence modeling and a Diffusion-based generative strategy with human-kinematic decomposition for high-quality motion synthesis. 
    
    \item Extensive experiments demonstrate that \textbf{BiTDiff} achieves state-of-the-art (SOTA) performance on \textbf{CM-Data}, and further supports joint-level motion editing for downstream human-AI interaction.
\end{itemize}

\begin{figure*}[t]
  \centering
\includegraphics[width=0.98\textwidth,trim=10 10 10 10,clip]{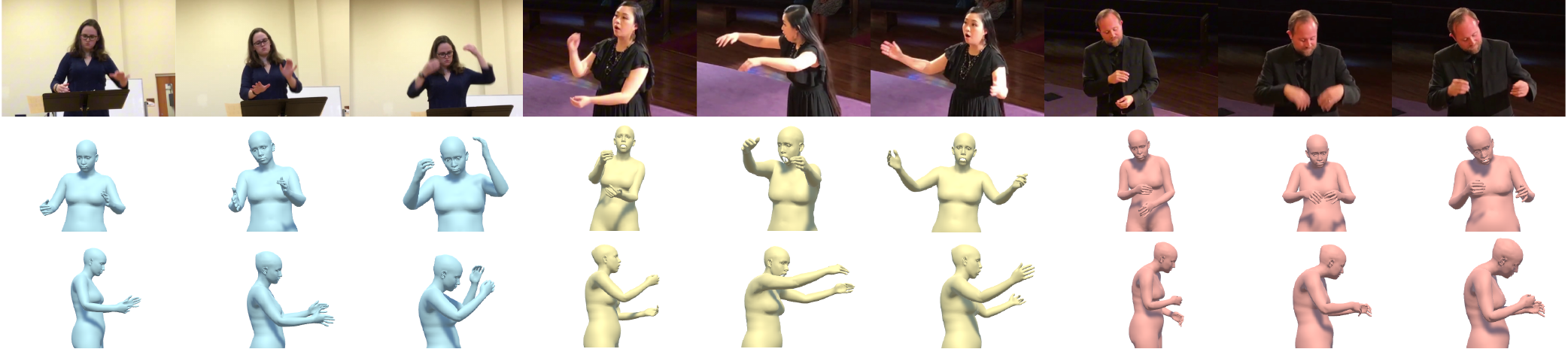}
  \vspace{-0.05in}
  \caption{\textbf{Examples from our pipeline.} Our reconstruction pipeline captures not only coarse body-level motions but also fine-grained details such as hand articulation and facial expressions. Moreover, multi-view visualizations demonstrate the accuracy and temporal stability of the recovered 3D conducting motions.}
  \label{fig:data}
  \vspace{-0.15in}
\end{figure*}

\section{Related Work}
\subsection{3D Gesture Generation}
Speech-driven 3D gesture generation aims to synthesize natural human gestures from speech and has made substantial progress in recent years. Existing methods can be broadly categorized into three families: (1) autoregressive-based, (2) diffusion-based, and (3) flow-matching-based approaches. 
\textbf{(1) Autoregressive-based methods.} These methods~\cite{yi2023generating,liu2025mag,liu2024emage} typically first construct discrete gesture units or motion tokens, followed by autoregressive modeling of speech-conditioned motion distributions over these units. Such designs are naturally suitable for streaming or real-time generation, but often have limited capacity in modeling complex and highly expressive motions. 
\textbf{(2) Diffusion-based methods.} These methods~\cite{zhang2025semtalk,zhang2025echomask,yang2024freetalker,zhang2024speech} substantially improve gesture realism, motion complexity, and diversity by modeling speech-driven gesture synthesis through iterative denoising, but usually come at the cost of high computational complexity during inference. 
\textbf{(3) Flow-matching-based methods.} More recent methods~\cite{zhang2026mitigating,liu2025gesturelsm} achieve generation quality comparable to diffusion-based approaches with only a few sampling steps, thereby further improving generation efficiency. 

However, these methods primarily focus on modeling speech-related semantic and prosodic information, while overlooking the music-structured control signals required in conducting motion generation, such as beat organization, ictus timing, and cueing.

\subsection{3D Dance Generation}
Music-to-dance generation has also achieved remarkable progress in recent years, particularly in the 3D setting. Existing methods can be broadly categorized into three families: (1) GAN-based, (2) autoregressive, and (3) diffusion-based approaches. 
\textbf{(1) GAN-based methods.} These methods~\cite{yang2024cohedancers,sun2019deep} synthesize dance motions from music through adversarial learning, where generators produce motions and discriminators provide supervision on realism and music--motion correspondence. While such methods improve motion fidelity to a certain extent, they still struggle to generate highly complex and compositionally rich dance movements. 
\textbf{(2) Autoregressive methods.} These methods~\cite{siyao2022bailando,siyao2023bailando++,yang2025megadance,yang2025matchdance,yang2026tokendance} first curate choreographic units, followed by autoregressive modeling of music-conditioned distributions over these units. This design enables long-horizon choreography modeling in a relatively cost-effective manner, but the generated motions often remain conservative due to the loss of the tokenization process. 
\textbf{(3) Diffusion-based methods.} These methods~\cite{tseng2023edge,li2023finedance,li2024lodge,li2024lodge++} corrupt motion sequences with noise and train denoising networks to iteratively recover dances conditioned on music, thereby jointly improving motion creativity, motion fidelity, and motion synchronization. However, these advantages usually come at the cost of substantially increased computational complexity during training and inference. 

However, these approaches mainly focus on body-level music-motion alignment, while overlooking the fine-grained hand, upper-body, and facial control signals required in conducting motion generation.

\subsection{Conducting Motion Generation}
Although this topic has received relatively limited attention, a few studies have explored 3D conducting motion generation. ~\cite{liu2022self} first introduced a large-scale open-source dataset based on 2D keypoints. However, this dataset is relatively coarse: on the one hand, it is mainly applicable to simple 2D settings; on the other hand, it focuses primarily on body movements while overlooking equally important facial and hand details. ~\cite{zhao2023taming} proposed a diffusion-based model, but due to the limitations of the dataset, the expressive capacity of 2D keypoints is inherently restricted, making it difficult to generalize to real-world industrial applications. Meanwhile, ~\cite{oh2024transfer} attempted to transfer the capability of recently popular 3D dance generation models to the conducting motion generation scenario. Although this approach enables 3D motion synthesis, it is still largely limited to body-level modeling, and its quality and efficiency remain unsatisfactory in long-sequence generation scenarios.

Overall, the field of 3D conducting motion generation currently faces two major challenges: (1) the lack of a large-scale and fine-grained dataset covering diverse conducting scenarios; and (2) the lack of effective methodology that can support long-sequence generation with high quality and efficiency.

\section{Dataset}

To tackle the dataset limitation, we develop a quality-oriented 3D conducting motion collection pipeline and construct \textbf{CM-Data}, a fine-grained 3D SMPL-X dataset with about 10 hours of conducting motion data. To the best of our knowledge, \textbf{CM-Data} is the first and largest public dataset for 3D conducting motion. Typical examples can be found in Fig.~\ref{fig:data}.

\subsection{Data Collection}
\subsubsection{3D-Friendly Internet Video Curation}
High-quality 3D conducting motion reconstruction depends critically on the visual quality, temporal continuity, and professionalism of the source videos. Therefore, rather than collecting large amounts of unconstrained Internet data, we adopt a quality-oriented curation strategy to select videos that are both suitable for fine-grained 3D motion recovery and representative of professional conducting practice. Specifically, we manually curate about 15 hours of conducting videos from online sources, focusing on professional performances such as conducting competitions and instructional demonstrations. We retain only videos that satisfy the following criteria: (1) stable camera viewpoints with limited shake or abrupt motion; (2) clear visibility of the conductor with minimal occlusion of key body parts, especially the arms, hands, and face; (3) limited shot changes and editing cuts to preserve temporal continuity; (4) clean lighting conditions and sufficient image resolution for reliable 3D reconstruction; and (5) clear conductor prominence over the background, with limited distraction from audiences, stage objects, or other performers.
These criteria improve the overall reliability of the subsequent 3D reconstruction process at the data-source level, and provide a cleaner foundation for fine-grained conducting motion modeling.

\subsubsection{Kinematic-Decomposition 3D Motion Reconstruction}
After obtaining curated conducting videos, we reconstruct fine-grained 3D conducting motions in the SMPL-X format through a kinematic-decomposition pipeline. Instead of relying on a single end-to-end model to recover all motion details, we decompose the reconstruction process into specialized subproblems corresponding to distinct kinematic components, namely the hand, face, and body. These components are subsequently aligned and fused into a unified SMPL-X representation, producing temporally coherent full-body motion sequences with detailed body, hand, and face dynamics.

Specifically, we use HaPTIC~\cite{ye2025predicting} for hand reconstruction, SPECTRE~\cite{filntisis2022visual} for face reconstruction, and PromptHMR~\cite{wang2025prompthmr} for body reconstruction:
\textbf{(1) Hand Reconstruction}
We adopt HaPTIC~\cite{ye2025predicting} for hand reconstruction because it directly models temporally coherent 4D hand motion from monocular videos, enabling more stable recovery of global hand trajectories than methods focused only on frame-wise 3D pose estimation. This makes it well suited for 3D conducting motion, where the conductor’s hands are the key medium for conveying rhythm, entrances, and expression, and thus require accurate reconstruction of both fine articulation and continuous motion dynamics.
\textbf{(2) Face Reconstruction}
We adopt SPECTRE~\cite{filntisis2022visual} for face reconstruction because it is a video-based 3D facial reconstruction method that focuses on perceptually faithful mouth and facial expression dynamics, especially through lipread-guided supervision of articulation-related movements. This is well suited for 3D conducting motion, where subtle facial expressions and mouth-related cues contribute importantly to musical expressiveness and thus require fine-grained dynamic reconstruction.
\textbf{(3) Body Reconstruction}
We adopt PromptHMR for body reconstruction because its promptable full-image design improves robustness to partial visibility and body truncation by leveraging scene context together with flexible spatial prompts such as face boxes, partial-body boxes, and masks. This is particularly suitable for 3D conducting videos, where the lower body is often partially visible or outside the frame, while the upper body, arm span, and torso coordination remain the dominant structural cues of conducting gestures.

\subsection{Data Statistic}
In total, \textbf{CM-Data} contains about 1{,}500 fine-grained 3D conducting motion samples, each with a duration ranging from 10 to 50 seconds, resulting in approximately 10 hours of SMPL-X motion data. To the best of our knowledge, \textbf{CM-Data} is the first and largest public dataset for 3D conducting motion generation. 

\textbf{CM-Data} offers several desirable properties for this task: (1) it covers a broad range of music-related performance scenarios, including orchestral conducting, choral conducting, vocal performance, and solo performance settings; (2) it spans diverse musical genres, including symphonic, operatic, choral, pop, and other contemporary music styles; (3) it is curated from multiple online platforms, including YouTube, TikTok, and Douyin, which increases the diversity of visual style and performance context; (4) it captures substantial variation in conducting style across different performers; (5) it provides fine-grained full-body, hand, and facial motion annotations, which are essential for modeling the expressive nature of conducting. Overall, these characteristics establish \textbf{CM-Data} as a strong benchmark for fine-grained 3D conducting motion generation.
For evaluation, we randomly select 60 samples to form the test set, while the remaining samples are used for training.

\begin{figure*}[!t]
  \centering
  \includegraphics[width=0.98\linewidth]{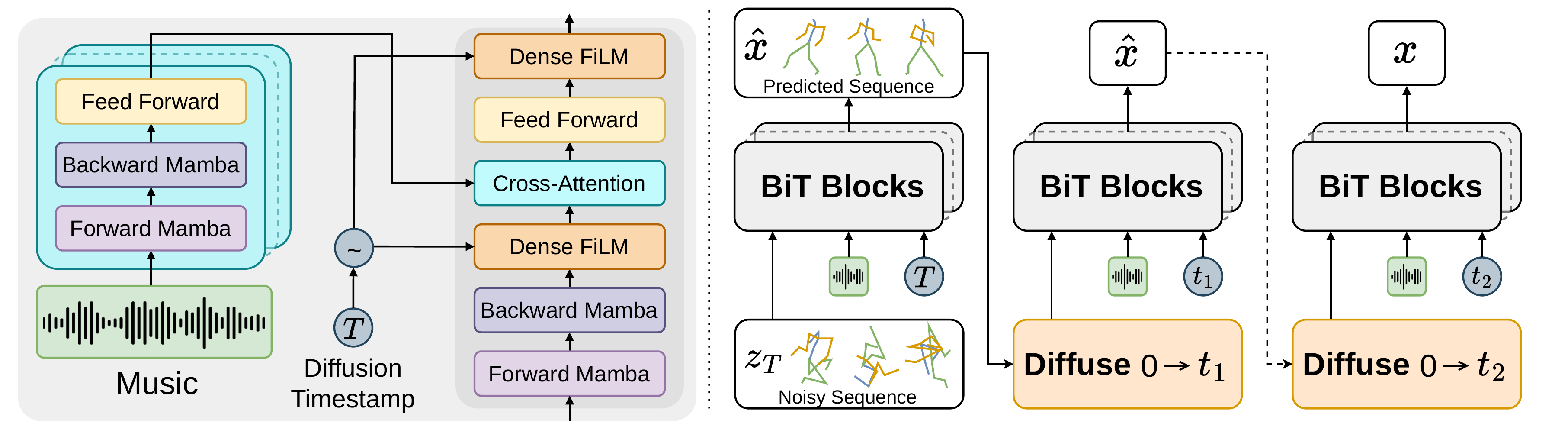}
  \vspace{-0.15in}
  \caption{\textbf{Overview of BiTDiff.} The left panel presents the detailed model architecture, while the right panel illustrates the inference strategy. Here, $0 < t_2 < t_1 < T$ indicate two intermediate timesteps in the diffusion process.}
  \label{fig:overview}
  \vspace{-0.15in}
\end{figure*}

\section{Methodology}
\subsection{Problem Definition}
Given a music sequence $M=\{m_0, m_1, \dots, m_T\}$, our goal is to generate a corresponding conducting motion sequence $C=\{c_0, c_1, \dots, c_T\}$. Each music feature $m_t$ is represented as a 35-dimensional vector extracted by Librosa~\cite{mcfee2015librosa}, including 20-dimensional MFCC, 12-dimensional Chroma, and three 1-dimensional features corresponding to Peak, Beat, and Envelope. Each conducting motion feature is represented as a 333-dimensional vector $c_t = [\tau; \theta]$, consisting of a 3-dimensional root translation $\tau$ and 330-dimensional 6D joint rotations~\cite{zhou2019continuity}. To ensure precise temporal correspondence between music and motion, we synchronize $M$ and $C$ at 30 FPS.

\subsection{Generative Strategy}
\subsubsection{Diffusion Model}
DDPM~\cite{ho2020denoising} defines diffusion as a Markov noising process with latents $\{z_t\}_{t=0}^T$ that follow a forward noising process $q(z_t|x)$, where $x \sim p(x)$ is drawn from the 3D conducting motion data distribution. The forward noising process is defined as:
\begin{equation}
q(z_t | x) \sim \mathcal{N}(\sqrt{\bar{\alpha}_t} x, (1 - \bar{\alpha}_t) I),
\end{equation}
where $\bar{\alpha}_t \in (0,1)$ are constants which follow a monotonically decreasing schedule such that when $\bar{\alpha}_t$ approaches 0. Timestep $T$ are commonly set to 1000, and $z_T \sim \mathcal{N}(0, I)$. With paired music conditioning $c$, we can reverse the forward diffusion process by learning to estimate $\hat{x}_\theta(z_t, t, c) \approx x$ with model parameters $\theta$ for all $t$ with condition $c$. We can optimize $\theta$ by the naive reconstruction loss in Diffusion Model~\cite{ho2020denoising}:
\begin{equation}
\mathcal{L}_{\text{rec}} = \mathbb{E}\big[\| \hat{x}_\theta(z_t, t, c) - x  \|_2^2 \big].
\end{equation}

\subsubsection{Training}
Since we adopt the 6D rotation representation~\cite{zhou2019continuity}, our motion parameterization does not suffer from the angular discontinuity issue. Therefore, $\mathcal{L}_{\text{rec}}$ can be directly applied to the SMPL-X face, body, and hand parameters. Beyond the reconstruction loss, auxiliary objectives are commonly introduced in kinematic motion generation to improve physical plausibility in the absence of explicit physical simulation~\cite{tseng2023edge}. 
Since the hands are located at the end of the human kinematic chain, joint losses computed through forward kinematics (FK) often underconstrain hand motions. To address this issue, we decompose the FK-based constraint into hand-specific and body-specific terms, which improves fine-grained hand modeling while preserving overall body coherence. Specifically, $\mathcal{L}_{\text{hand}}$ is computed by retaining only the hand-related components $\hat{x_{\text{h}}},x_{\text{h}}$ in SMPL-X while setting the body-related components to zero, whereas $\mathcal{L}_{\text{body}}$ is computed by retaining only the body-related components $\hat{x_{\text{b}}},x_{\text{b}}$ while zeroing out the hand-related components. 
To further enhance motion smoothness and strengthen the model’s ability to capture temporal dynamics, we additionally introduce a velocity loss.
\begin{equation}
\begin{aligned}
\mathcal{L}_{\text{hand}} &= \mathbb{E}\big[ \| (FK(\hat{x_{\text{h}}}) - FK(x_{\text{h}}))
+
(FK(\hat{x_{\text{h}}})' - FK(x_{\text{h}})')
\|_2^2 \big], \\
\mathcal{L}_{\text{body}} &= \mathbb{E}\big[ \| (FK(\hat{x_{\text{b}}}) - FK(x_{\text{b}}))
+
(FK(\hat{x_{\text{b}}})' - FK(x_{\text{b}})')
\|_2^2 \big] \\
\mathcal{L}_{\text{foot}} &= \mathbb{E}\big[\| FK(\hat{x})' \cdot \hat{\mathbf{b}}  \|_2^2 \big], \\
\end{aligned}
\end{equation}
where $FK(\cdot)$ denotes the forward kinematic function that converts joint angles into joint positions,  and $\hat{\mathbf{b}}$ is the model's own prediction of the binary foot contact label's portion of the pose. Our overall training loss $\mathcal{L}$ is the weighted sum of the above losses, where the weights $\lambda$ were chosen to balance the magnitudes of the losses:
\begin{equation}
\mathcal{L} = 
\lambda_{\text{rec}} \mathcal{L}_{\text{rec}} + \lambda_{\text{hand}} \mathcal{L}_{\text{hand}} + \lambda_{\text{body}} \mathcal{L}_{\text{body}} + \lambda_{\text{foot}} \mathcal{L}_{\text{foot}}.
\end{equation}

\subsubsection{Inference}
At each of the denoising timesteps $t$, BiTDiff predicts the denoised sample and noises it back to timestep $t-1$:
$\hat{z}_{t-1} \sim q(\hat{x}_{\theta}(\hat{z}_t, c), t-1)$, terminating when it reaches $t=0$. 
If a DDIM-style sampling strategy is adopted, the model can directly move from timestep $t$ to an arbitrary earlier timestep $t_1$, rather than only to $t-1$, as illustrated in Fig.~\ref{fig:overview}.
We train our model using classifier-free guidance (CFG), which is commonly used in diffusion-based models.
Following ~\cite{tseng2023edge}, we implement CFG by randomly replacing the conditioning with $c=\emptyset$ during training with low probability (e.g., 20\%).
Guided inference is then expressed as the weighted sum of unconditionally and conditionally generated samples. At sampling time, we can amplify the conditioning $c$ by choosing a guidance weight $w > 1$:
\begin{equation}
\tilde{x}(\hat{z}_t, c) =
\hat{x}(\hat{z}_t, \emptyset) + w \cdot (\hat{x}(\hat{z}_t, c) - \hat{x}(\hat{z}_t, \emptyset)).
\end{equation}

\subsubsection{Motion Editing}
To enable editing for conducting motions generated by BiTDiff, we adopt the standard \textit{masked denoising} technique. Let the conducting motion sequence be $x \in \mathbb{R}^{T \times D}$, where $T$ is the sequence length and $D$ is the motion dimension. Given a partial constraint $x^{\text{known}}$ and a binary mask $m \in \{0,1\}^{T \times D}$ indicating the constrained entries, we perform the following replacement at each denoising timestep:
\begin{equation}
\hat{z}_{t-1} := m \odot q(x^{\text{known}}, t-1) + (1 - m) \odot \hat{z}_{t-1},
\end{equation}
where $\odot$ denotes the Hadamard product. In this way, the constrained entries are fixed by the user, while the remaining entries are generated by the model. Since $m$ can be defined over temporal regions, joint subsets, or both, this formulation naturally supports flexible motion editing at inference time without additional training, as follows:
\textbf{(1) Temporal in-betweening.}
Let $\mathcal{T}_{\text{past}}$ and $\mathcal{T}_{\text{future}}$ denote the known prefix and suffix time intervals, respectively. We define $m_{t,d}=1$ for all $t \in \mathcal{T}_{\text{past}} \cup \mathcal{T}_{\text{future}}$ and all motion dimensions $d$, while the middle interval remains unconstrained. BiTDiff then inpaints the missing segment with smooth transitions and coherent conducting dynamics, which is useful for motion refinement and sparse key-segment-based authoring.
\textbf{(2) Temporal continuation / streaming generation.}
Let $\mathcal{T}_{\text{obs}}$ denote the observed prefix interval. We set $m_{t,d}=1$ for all $t \in \mathcal{T}_{\text{obs}}$ and all $d$, while leaving future timesteps unconstrained. BiTDiff can then progressively generate the subsequent motion in a chunk-wise manner while maintaining temporal stability and musical consistency, making it suitable for low-latency streaming generation and real-time human-AI conducting interaction.
\textbf{(3) Upper-to-lower body completion.}
Let $\mathcal{J}_{\mathrm{up}}$ and $\mathcal{J}_{\mathrm{low}}$ denote the upper-body and lower-body joint sets, respectively. We define $m_{t,d}=1$ for motion dimensions $d$ associated with $\mathcal{J}_{\mathrm{up}}$ at all timesteps $t$, while dimensions corresponding to $\mathcal{J}_{\mathrm{low}}$ remain unconstrained. BiTDiff can thus synthesize plausible lower-body motion coordinated with the given conducting dynamics, which is useful for partial-body animation completion and controllable motion design.
\textbf{(4) Body-to-hand/face enrichment.}
Let $\mathcal{J}_{\mathrm{body}}$, $\mathcal{J}_{\mathrm{hand}}$, and $\mathcal{J}_{\mathrm{face}}$ denote the body, hand, and face components in SMPL-X. We set $m_{t,d}=1$ for dimensions $d$ associated with $\mathcal{J}_{\mathrm{body}}$ at all timesteps $t$, while leaving those associated with $\mathcal{J}_{\mathrm{hand}} \cup \mathcal{J}_{\mathrm{face}}$ unconstrained. BiTDiff can then synthesize detailed hand and facial dynamics consistent with the global conducting pattern, enabling fine-grained expressive enrichment for digital human animation and conducting authoring.

\subsection{Model Architecture}
\subsubsection{Overview}
BiTDiff adopts a BiMamba–Transformer hybrid model architecture, thereby enabling the generation of temporally coherent and musically aligned conducting motions. BiMamba captures intra-modal dependencies in music or dance, while the Transformer models cross-modal context. As shown in Fig. ~\ref{fig:overview}, the architecture details are as follows:
Firstly, our model conditions the generator on the Librosa~\cite{mcfee2015librosa}-extracted music features as~\cite{li2021ai}, which are then processed by an $L_m$‑layer BiMamba to capture intra‑modal temporal dynamics. 
Secondly, the diffusion time step $t$ is encoded as sinusoidal embeddings and fused by element-wise addition to yield a timestep embedding. 
Thirdly, the motion generator consists of $L_c$ stacked blocks. In each block: (1) the current state $z_t$ is first passed through a BiMamba to model intra-modal local dependencies; (2) FiLM ~\cite{perez2018film} is applied to modulate the features with the timestep embedding; (3) a Transformer performs cross-modal attention over the music encoding to integrate global musical context, and subsequently passes the result through a feed-forward network; and (4) a second FiLM~\cite{perez2018film} further reinforces the timestep conditioning. Finally, the generator outputs the 3D motion sequence $\hat{x}_\theta(z_t, t, c)$, represented as SMPL-X parameters.

\subsubsection{Long-sequence Generation}
Because BiMamba serves as the primary temporal backbone, BiTDiff inherits strong long-range modeling capacity and can be naturally extended from short-sequence training to long-sequence generation at inference time in the same non-autoregressive manner, without relying on autoregressive rollout or segment-wise stitching. This generation paradigm naturally avoids the exposure-bias accumulation in autoregressive methods~\cite{yang2025megadance,siyao2023bailando++,siyao2024duolando} as well as the unstable transition regions commonly introduced by inpainting-based methods~\cite{tseng2023edge,li2024lodge}. Moreover, BiTDiff can support both one-shot long-sequence generation and online streaming generation with low latency, making BiTDiff well suited for interactive human-AI conducting applications.

\subsubsection{Intra-Modal BiMamba}
While the Transformer is powerful for modeling global dependencies, it is inherently position-invariant and captures sequence order mainly through positional encodings~\cite{vaswani2017attention}, which limits its ability to model fine-grained local temporal continuity. In contrast, music-driven conducting motion generation requires strong local coherence between adjacent movements. Owing to its inherent sequential inductive bias, Mamba~\cite{gu2023mamba} has demonstrated strong capability in modeling fine-grained local dependencies~\cite{xu2024mambatalk}. Moreover, its linear computational complexity provides a clear efficiency advantage in long-sequence settings. Building upon this, Bidirectional Mamba processes inputs in both forward and backward directions, enabling richer contextual representations and a deeper understanding of music and motion.
Specifically, the Selective State Space Model (Mamba) integrates a selection mechanism and a scan module (S6)~\cite{gu2023mamba} to dynamically emphasize salient input segments for efficient sequence modeling. Unlike traditional SSMs with time-invariant parameters, Mamba generates input-dependent $\bar{A}_t, \bar{B}_t, C_t$ through fully connected layers, enhancing generalization. For each time step $t$, the input $x_t$, hidden state $h_t$, and output $y_t$ evolve as:
\begin{equation}
h_t = \bar{A}_t h_{t-1} + \bar{B}_t x_t,\quad y_t = C_t h_t,
\end{equation}
where $\bar{A}_t, \bar{B}_t, C_t$ are dynamically updated, and the state transitions become:
\begin{equation}
\bar{A} = \exp(\Delta A),\quad
\bar{B} = (\Delta A)^{-1}(\exp(\Delta A)-I)\cdot \Delta B,
\end{equation}
where $\Delta$ is the discretization step size, $A$ is the continuous-time state transition matrix, $B$ is the input projection matrix, and $C$ is the output projection matrix.

\subsubsection{Cross-Modal Transformer}
While BiMamba~\cite{gu2023mamba} is effective at modeling intra-modal local dependencies, conducting motion generation also requires cross-modal alignment between motion and music at a more global semantic level, such as musical phrasing, dynamic progression, and beat structure. To capture such complementary global context, we introduce a Transformer~\cite{vaswani2017attention} block for cross-modal interaction. Specifically, the current conducting motion features are used as queries $Q_c$, while the encoded music features serve as keys $K_m$ and values $V_m$, allowing the model to selectively attend to the most relevant musical cues during motion generation. This block consists of a cross-attention layer followed by a feed-forward network (FFN), where the former retrieves cross-modal information and the latter further refines the fused representation. The attention layer is formulated as:
\begin{equation}
\text{Attention}(Q_c, K_m, V_m) = \text{Softmax}\left(\frac{Q_c \cdot K_m^T}{\sqrt{C}}\right) \cdot V_m.
\end{equation}

\begin{table*}[!t]
\centering
\renewcommand{\arraystretch}{1.2}
\setlength{\tabcolsep}{5pt}
\caption{Quantitative comparison of generation quality and efficiency on the \textbf{CM-Data} dataset.}
\vspace{-0.1in}
\label{tab:quantitative}
\scriptsize
\resizebox{0.97\linewidth}{!}{
\begin{tabular}{l|cc|cc|cc|c|cc}
\toprule
& \multicolumn{2}{c|}{\textbf{Hand}}
& \multicolumn{2}{c|}{\textbf{Body}}
& \multicolumn{2}{c|}{\textbf{Face}}
& \textbf{Alignment}
& \multicolumn{2}{c}{\textbf{Efficiency}} \\
\cmidrule(l{-0.5pt}r{0pt}){2-3}
\cmidrule(l{0pt}r{0pt}){4-5}
\cmidrule(l{0pt}r{0pt}){6-7}
\cmidrule(l{0pt}r{0pt}){8-8}
\cmidrule(l{0pt}r{0pt}){9-10}
& FID$_{\text{h}}$$\downarrow$ & DIV$_{\text{h}}$$\uparrow$
& FID$_{\text{b}}$$\downarrow$ & DIV$_{\text{b}}$$\uparrow$
& FID$_{\text{f}}$$\downarrow$ & DIV$_{\text{f}}$$\uparrow$
& BAS$\uparrow$
& L@1024$\downarrow$ & L@4096$\downarrow$ \\
\midrule
Ground Truth & -- & 10.69 & -- & 3.01 & -- & 5.60 & 0.272 & -- & -- \\
VirtualConductor~\cite{liu2022self} & 101.57 & 6.42 & 86.83 & 2.31 & 48.64 & 1.41 & 0.211 & 3.84s & 10.92s \\
DiffusionConductor~\cite{zhao2023taming} & 39.42 & 8.91 & 31.76 & 2.84 & 42.18 & 3.08 & 0.286 & 18.73s & 74.12s \\
EDGE~\cite{tseng2023edge} & 36.28 & 9.14 & 27.95 & 2.96 & 39.87 & 3.26 & 0.289 & 5.41 & 20.13 \\
Lodge~\cite{li2024lodge} & 37.11 & 10.02 & 20.96 & 3.54 & 30.85 & \textbf{3.35} & 0.296 & 3.62s & 9.51s \\
MambaTalk~\cite{xu2024mambatalk} & 53.74 & 7.95 & 41.62 & 2.47 & 52.39 & 2.76 & 0.254 & 2.27s & 3.02s \\
DiffSHEG~\cite{chen2024diffsheg} & 27.43 & 9.36 & 29.08 & \textbf{3.88} & 38.76 & 3.24 & 0.284 & 17.53s & 73.20s \\
\midrule
BiTDiff (\textbf{Ours}) & \textbf{25.81} & \textbf{10.34} & \textbf{19.14} & 3.78 & \textbf{27.89} & 3.22 & \textbf{0.302} & \textbf{1.44 s} & \textbf{2.56 s} \\
\bottomrule
\end{tabular}
}
\vspace{-0.1in}
\end{table*}

\section{Experiment}
\subsection{Comparison}
\subsubsection{Generation Quality}
As the first study on fine-grained 3D conducting motion generation, we compare \textbf{BiTDiff} against three groups of representative baselines: (1) \textit{2D conducting motion generation} methods, including DiffusionConductor~\cite{zhao2023taming} and VirtualConductor~\cite{liu2022self}; (2) \textit{3D dance generation} methods, including EDGE~\cite{tseng2023edge}, and Lodge~\cite{li2024lodge}; and (3) \textit{3D gesture generation} methods, including MambaTalk~\cite{xu2024mambatalk} and DiffSHEG~\cite{chen2024diffsheg}. Since all baselines need to be retrained on our dataset, we select well-documented open-source methods that are representative and influential in their respective fields, although they are not necessarily the most recent ones.
For evaluation, we separately extract kinetic features~\cite{li2023finedance} for the face, hand, and body, and compute FID and DIV to measure motion fidelity and diversity, respectively. For motion-music synchronization, we follow prior work~\cite{li2021ai} and adopt Beat Alignment Similarity (BAS) based on SMPL-X keypoints. We additionally exclude MSE and MAE, since music-driven conducting generation is inherently one-to-many: for the face, hand, and body, a given music input may correspond to multiple plausible motion realizations. In contrast, tasks such as speech-driven gesture generation often involve stronger correspondence between facial motion and speech, while video-driven pose estimation imposes more direct constraints on hand and body poses from the input video.

As shown in Table~\ref{tab:quantitative}, \textbf{BiTDiff} consistently outperforms all baseline methods across the three evaluation dimensions, achieving the best overall generation quality. In particular, compared with the strongest baseline, \textbf{Lodge}, \textbf{BiTDiff} reduces the average \textbf{FID} by \textbf{18.1}, improves the average \textbf{DIV} by \textbf{2.5}, and increases \textbf{BAS} by \textbf{2.0}, demonstrating clear advantages in motion fidelity, diversity, and music-motion synchronization. These results indicate that \textbf{BiTDiff} establishes a new state of the art for fine-grained 3D conducting motion generation. We attribute this superiority to two key factors: (1) the diffusion-based generative strategy provides strong capacity for modeling complex and diverse motion distributions; and (2) the proposed \textbf{BiMamba-Transformer} hybrid architecture effectively captures both local temporal dynamics and global music-motion dependencies, leading to more realistic, expressive, and synchronized conducting motions.

\subsubsection{Generation Efficiency}
We evaluate generation efficiency by measuring the latency of generating 1,024-frame and 4,096-frame motion sequences (Latency@1024 and Latency@4096) on an NVIDIA H20 GPU. As shown in Table~\ref{tab:quantitative}, \textbf{BiTDiff} achieves the best efficiency among all compared methods, and its advantage becomes more pronounced for long-sequence generation. Compared with the second-fastest baseline (\textbf{MambaTalk}), \textbf{BiTDiff} further reduces latency by \textbf{36.6\%} at 1,024 frames and \textbf{15.2\%} at 4,096 frames. This verifies the superior efficiency of \textbf{BiTDiff} in practical deployment scenarios. The improvement mainly comes from the proposed BiMamba-Transformer hybrid architecture, which supports memory-efficient non-autoregressive generation and thus enables scalable long-horizon motion synthesis.

\subsubsection{Qualitative Analysis}
\begin{figure}[!t]
  \centering
  \includegraphics[width=0.98\linewidth]{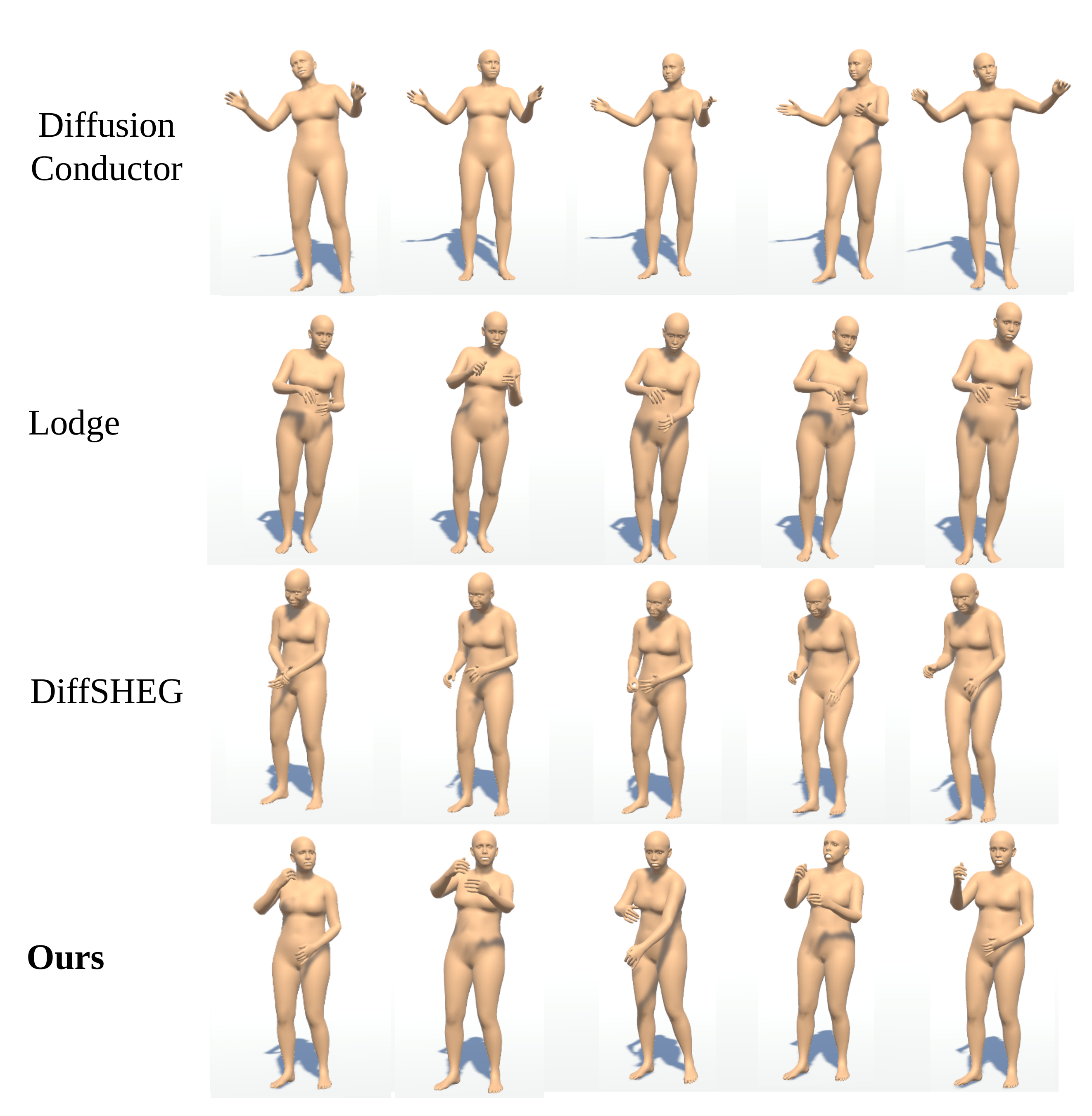}
  \vspace{-0.05in}
  \caption{Qualitative comparison with SOTAs.}
  \vspace{-0.1in}
  \label{fig:cmp}  
\end{figure}

As shown in Fig.~\ref{fig:cmp}, \textbf{BiTDiff} produces conducting motions that are noticeably more expressive and temporally stable. In particular, \textbf{BiTDiff} better captures fine-grained variations in gesture amplitude, hand articulation, upper-body coordination, and facial dynamics, while also generating motions that are more diverse and creatively varied. By contrast, the motions generated by other methods are relatively monotonous and less stable over time. Specifically, \textbf{Lodge~\cite{li2024lodge}} often produces movements that are less consistent with realistic conducting gestures, \textbf{DiffSHEG~\cite{chen2024diffsheg}} tends to lose facial expressiveness and fine-grained hand motion details, and \textbf{DiffusionConductor~\cite{zhao2023taming}} frequently generates repetitive motion patterns.

\subsection{User Study}
\subsubsection{Experimental Setup}
User feedback is essential for evaluating generation quality in the music-driven conducting motion generation task, due to its inherent subjectivity\cite{legrand2009perceiving}. Following ~\cite{yang2025megadance}, we randomly select 30 real-world music segments, each lasting 30 seconds, and generated motion sequences using the models described above. These sequences are evaluated through a double-blind questionnaire completed by 30 participants with conducting backgrounds. Participants are compensated at a rate exceeding the local average hourly wage. The questionnaires used a 5-point scale (Great, Good, Fair, Bad, Terrible) to assess three aspects: Motion Synchronization (MS, alignment with rhythm and style), Motion Fidelity (MF, physical plausibility), and Motion Creativity (MC, diversity and complexity). We additionally include catch trials with ground-truth and distorted-motion videos. Participants who fail to assign higher scores to the ground-truth videos and lower scores to the distorted ones are excluded from the final evaluation.

\subsubsection{Result Analysis}
As shown in Tab.~\ref{tab:user_study}, \textbf{BiTDiff} achieves the best overall user ratings among all compared methods across the three evaluation aspects, with the most notable advantage in \textbf{Motion Creativity} (4.17). This indicates that our method is more capable of generating conducting motions that are not only physically plausible and well aligned with the music, but also more diverse and compositionally rich from the perspective of human perception. Although a gap still remains between generated results and \textit{Ground Truth}, \textbf{BiTDiff} already attains strong subjective performance, suggesting that it can produce high-quality conducting motions that are favorably perceived by human evaluators. Overall, these results demonstrate the superiority of \textbf{BiTDiff} under human preference-based evaluation, and also validate the effectiveness of \textbf{CM-Data} as a high-quality benchmark that can support meaningful research on music-driven conducting motion generation.

\subsection{Ablation}
\subsubsection{Generative Strategy}
We conduct ablation studies on the proposed generative strategy from two aspects: (1) removing the velocity loss, denoted as \textit{w/o Vel.}, and (2) replacing the proposed hand/body kinematic decomposition with a naive unified FK loss, denoted as \textit{Naive FK}. As shown in Table~\ref{tab:ablation}, removing the velocity loss leads to a slight degradation in body-related fidelity, indicating that temporal smoothness is important for stabilizing motion transitions and improving the realism of generated conducting motions. In addition, replacing the proposed kinematic decomposition with a naive FK loss causes a clear deterioration on hand-related metrics, especially Hand FID and DIV, while the other metrics remain largely unchanged. This suggests that directly applying a unified FK constraint is insufficient for fine-grained hand supervision, since hand motions are located at the end of the human kinematic chain and are more difficult to constrain effectively.

\subsubsection{Model Architecture}
We further evaluate the proposed architecture using two variants. First, we replace the intra-modal \textbf{BiMamba} with a standard one-directional \textbf{Mamba}. Second, we replace \textbf{BiMamba} with a pure \textbf{Transformer} backbone. Since pure Transformer modeling performs poorly under our non-autoregressive setting and tends to fall into poor local minima, we adopt a progressive inpainting strategy similar to EDGE~\cite{tseng2023edge} for this variant. As shown in Table~\ref{tab:ablation}, replacing BiMamba with Mamba slightly improves generation efficiency, but causes a clear drop in generation quality, highlighting the importance of bidirectional temporal modeling for fine-grained conducting motion generation. In contrast, the Transformer-based variant achieves generation quality close to the full model, with only a slight overall decrease. However, its progressive generation process introduces a large efficiency overhead, making it unsuitable for real-time human-AI interaction. Overall, these results show that the proposed BiMamba-Transformer hybrid achieves the best balance between generation quality and efficiency, which is exactly the design goal of \textbf{BiTDiff}.

\subsection{Motion Editing}
\begin{figure}[!t]
  \centering
  \includegraphics[width=0.8\linewidth]{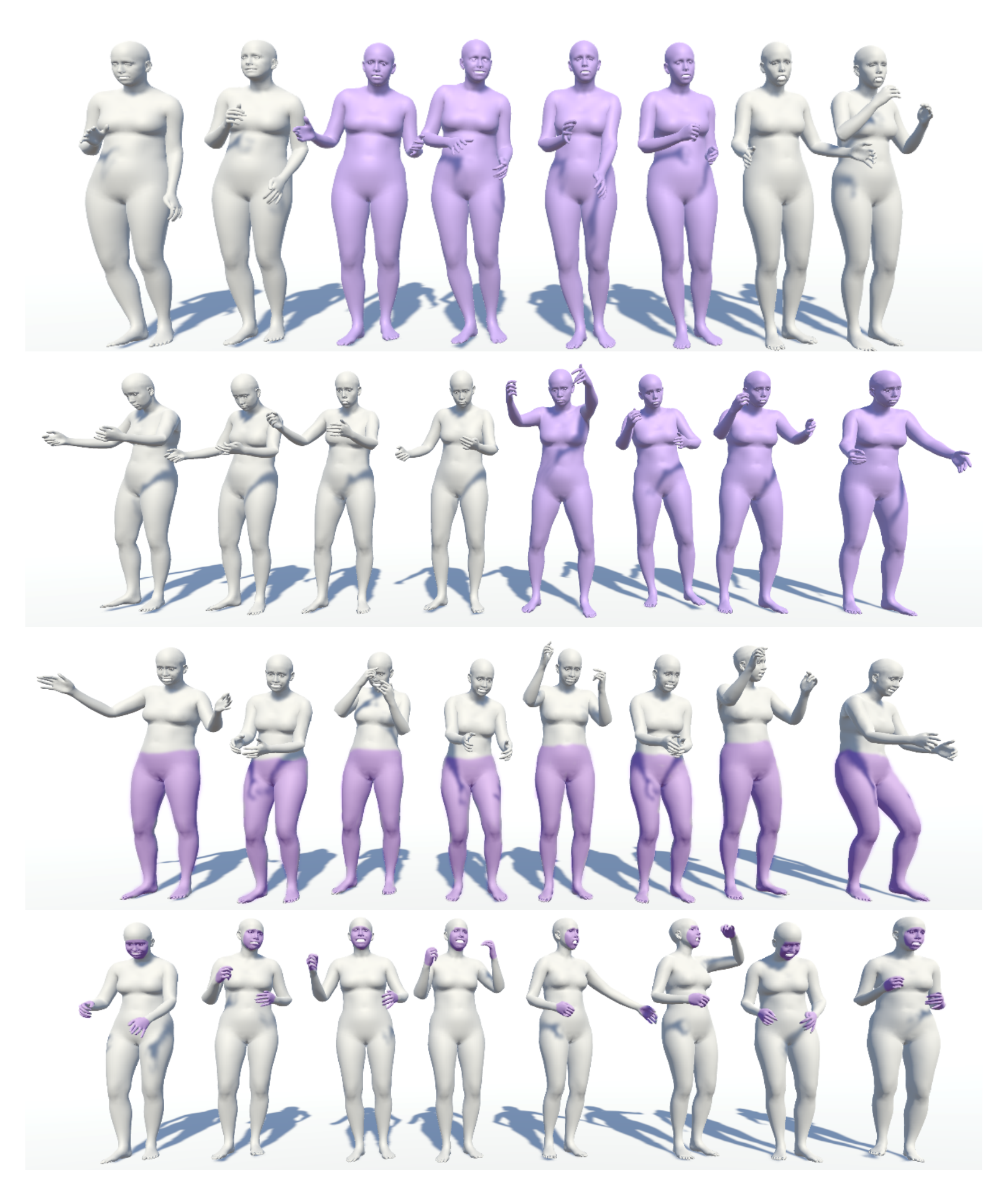}
  \vspace{-0.15in}
  \caption{Motion editing visualization of BiTDiff.}
  \vspace{-0.15in}
  \label{fig:me}  
\end{figure}

As shown in Fig.~\ref{fig:me}, \textbf{BiTDiff} supports flexible and effective motion editing at both the temporal and joint levels, enabling controllable conducting motion generation under partial constraints.

\subsubsection{Temporal-Level}
BiTDiff supports temporally constrained editing through masked denoising. (1) Given both preceding and following motion segments, it can plausibly inpaint the missing interval with smooth transitions and coherent conducting dynamics, which is useful for motion refinement and sparse key-segment-based authoring. (2) Given only the preceding segment, BiTDiff can progressively generate subsequent motion in a chunk-wise manner while maintaining temporal stability and musical consistency, making it suitable for low-latency streaming generation and real-time human-AI conducting interaction.

\subsubsection{Joint-Level}
BiTDiff also enables joint-level editing under partial body constraints. (1) Given the upper-body motion, it can generate plausible lower-body movements that remain coordinated with the conducting dynamics, which is useful for partial-body animation completion and controllable motion design. (2) Given only body motion, BiTDiff can further synthesize detailed hand and facial dynamics that match the global conducting pattern, supporting fine-grained motion enrichment for digital human animation and expressive conducting authoring.

\begin{table}[!t]
  \centering
  \caption{User study on the \textbf{CM-Data} dataset.}
  \vspace{-0.1in}
  \resizebox{0.3\textwidth}{!}{
    \begin{tabular}{lccc}
      \toprule
      Method & MS $\uparrow$ & MF $\uparrow$ & MC $\uparrow$ \\
      \midrule
      Ground Truth 
        & 4.21 & 4.47 & 4.09 \\
      EDGE~\cite{tseng2023edge} 
        & 3.41 & 3.35 & 3.28 \\
      DiffSHEG~\cite{chen2024diffsheg} 
        & 3.46 & 3.38 & 3.31 \\
      Lodge~\cite{li2024lodge} 
        & 3.49 & 3.43 & 3.36 \\
      BiTDiff (\textbf{Ours}) 
        & \textbf{3.96} & \textbf{3.88} & \textbf{4.17} \\
      \bottomrule
    \end{tabular}
  }
  \label{tab:user_study}
  \vspace{-0.15in}
\end{table}

\begin{table}[!t]
\centering
\caption{Ablation study on \textbf{CM-Data}.}
\vspace{-0.1in}
\resizebox{0.48\textwidth}{!}{
\begin{tabular}{lcccccc}
\toprule
Method 
& FID$_h$$\downarrow$ 
& DIV$_h$$\uparrow$
& FID$_b$$\downarrow$ 
& DIV$_b$$\uparrow$
& BAS$\uparrow$
& L@4096$\downarrow$ \\
\midrule

w/o Vel.      
& 30.47 & 9.88 & 22.63 & 3.93 & 0.247 & 2.56s \\
Naive FK      
& 69.38 & 6.72  & \textbf{17.58} & \textbf{4.15} & 0.265 & 2.56s \\

Mamba (uni)   
& 47.96 & 8.68  & 31.41 & 3.74 & 0.281 & \textbf{1.91s} \\
Transformer   
& 27.33 & 10.02 & 18.87 & 3.71 & 0.298 & 13.84s \\

\midrule
BiTDiff (Full) 
& \textbf{25.81} & \textbf{10.34} 
& 19.14 & 3.78 
& \textbf{0.302} 
& 2.56s \\
\bottomrule
\end{tabular}
}
\label{tab:ablation}
\vspace{-0.15in}
\end{table}

\section{Conclusion}
In this paper, we investigate the underexplored task of fine-grained 3D conducting motion generation. To address the lack of suitable data in this field, we develop a quality-oriented 3D conducting motion collection pipeline and construct \textbf{CM-Data}, a large-scale fine-grained 3D SMPL-X dataset for conducting motion generation. To address the methodological challenge of jointly achieving high quality and high efficiency long-sequence generation, we further propose \textbf{BiTDiff}, a novel framework built upon a diffusion-based generative strategy and a BiMamba-Transformer hybrid architecture. Extensive experiments demonstrate that \textbf{BiTDiff} achieves state-of-the-art performance on \textbf{CM-Data}, and further supports joint-level motion editing for downstream human-AI interaction.

We hope that \textbf{CM-Data} and \textbf{BiTDiff} can provide a strong foundation for future research on conducting motion understanding and generation. In future work, we aim to incorporate text-based control to provide users with more flexible and intuitive motion guidance, and to design more effective downstream human-AI interaction systems for practical deployment.

\bibliographystyle{ACM-Reference-Format}
\balance
\bibliography{mm-paper}

\end{document}


\title{BiTDiff: Fine-Grained 3D Conducting Motion Generation \\ via BiMamba-Transformer Diffusion}

\author{Supplementary Material}

\renewcommand{\shortauthors}{Tianzhi Jia et al.}

\begin{abstract}
This appendix contains additional materials for the paper “BiTDiff: Fine-Grained 3D Conducting Motion Generation via BiMamba-Transformer Diffusion”. The appendix is organized as follows: 
    \begin{itemize}
      \item Implementation Details.
      \item Qualitative Analysis.
      \item Dataset Details.  
      \item User Study Questionnaire.
      \item Limitation and Future Work
    \end{itemize}
\end{abstract}

\ccsdesc[500]{Applied computing~Arts and humanities}
\ccsdesc[500]{Human-centered computing}
\ccsdesc[500]{Computing methodologies~Computer vision}
\ccsdesc[500]{Computing methodologies~Animation}

\keywords{AI for Art, AI Generative Content, Digital Human, 3D Motion Generation, Music-Driven Conducting Motion Generation}

\maketitle

\section{Implementation Details}
\subsection{Training Setup}
We implement \textbf{BiTDiff} based on a diffusion model with a BiMamba-Transformer hybrid architecture and train it on the proposed \textbf{CM-Data} dataset. All experiments are conducted using the \texttt{Accelerate} library for distributed training on 8 NVIDIA H20 Tensor Core GPUs. We use a global batch size of 128 and train the model for 3000 epochs. The random seed is fixed to 42 for all experiments.
For optimization, we adopt the Adam optimizer with a learning rate of $4 \times 10^{-4}$ and a weight decay of $0.02$. To improve training stability, we maintain an exponential moving average (EMA) of model parameters with a decay rate of $0.9999$. Model checkpoints are periodically saved every 50 epochs for evaluation.

\subsection{Data Representation and Preprocessing}
We follow the motion representation described in the main paper. Specifically, each motion frame is represented as a 333-dimensional vector consisting of root translation and 6D joint rotations in the SMPL-X parameterization, while the music condition is represented by the 35-dimensional Librosa feature described in the main paper. Music and motion are temporally aligned at 30 FPS.
Before training, motion features are standardized. For rotational parameters, we consistently use the 6D rotation representation. During training, we extract motion clips using a sliding-window strategy. Unless otherwise specified, the model is trained on motion sequences of 240 frames (8 seconds at 30 FPS).

\subsection{Diffusion and Training Objective}
We adopt a standard denoising diffusion probabilistic model with a total of 1000 diffusion timesteps. During training, we use a linear noise schedule. The model is optimized with a weighted multi-objective loss composed of reconstruction loss, 3D hand joint position loss, 3D body joint position loss, and foot contact loss. The corresponding loss weights are set to
$\lambda_{\text{rec}}{=}0.636$,
$\lambda_{\text{hand}}{=}0.636$,
$\lambda_{\text{body}}{=}0.636$, and
$\lambda_{\text{foot}}{=}10.942$.
The reconstruction term supervises the denoised motion directly in the motion parameter space, while the hand and body terms provide additional forward-kinematics-based geometric constraints for different kinematic components. The foot contact term is further included to suppress implausible foot sliding and improve physical consistency during generation.
For classifier-free guidance training, we randomly drop the condition with a probability of 20\%, following the standard unconditional-conditioning strategy.

\subsection{Model Architecture}
The conditional encoding branch contains 2 layers of BiMamba, and the motion generation branch consists of 8 stacked BiMamba-Transformer blocks. The latent feature dimension is set to 512 throughout the network.
For each Mamba unit, we set the state size to 16, the convolution kernel size to 4, and the expansion factor to 2. For each Transformer block, we use 4 attention heads, a feed-forward hidden dimension of 1024, a dropout rate of 0.1, and GELU as the activation function.
Overall, this hybrid design enables the model to capture efficient intra-modal temporal dynamics through BiMamba while incorporating cross-modal global music context through Transformer-based attention.

\subsection{Inference Setup}
At inference time, we use DDIM sampling with 50 denoising steps. For classifier-free guidance, the guidance weight is set to $w{=}4$. Unless otherwise specified, all quantitative and qualitative results of \textbf{BiTDiff} are obtained under this setting.
To evaluate long-horizon generation ability, we perform one-shot inference on sequences of 1024 frames (34.13 seconds), rather than chunk-wise generation. This setting is intentionally more challenging than the training setting and is used to evaluate the scalability of \textbf{BiTDiff} for long-sequence conducting motion generation.
For latency comparison, we only measure the GPU sequence-generation time. The reported latency does not include music feature extraction, motion post-processing, or rendering time.

\section{Qualitative Analysis}
In this section, we provide additional qualitative analysis for both the comparison results and the ablation study. Since conducting motion generation is a strongly temporal task, static images are often insufficient to fully reflect the quality of the generated motions, especially in terms of temporal coherence, rhythmic variation, expressive development, and fine-grained motion details. Therefore, we encourage readers to refer directly to the supplementary video, where the qualitative differences can be observed more clearly and comprehensively.

\subsection{Comparison}
To complement the quantitative results in the main paper, we provide additional qualitative comparisons in the supplementary materials. Following the presentation style commonly used in prior music-driven motion generation work, we organize the comparison into two parts. In the first part, we present three pairwise \textit{1v1} comparisons between \textbf{BiTDiff} and representative baselines. In the second part, we further present two \textit{1v1v1v1} comparisons to provide a more direct side-by-side comparison across multiple methods under the same music input. Whenever possible, we select representative cases directly from model predictions; only when a sufficiently clear case cannot be found do we additionally refer to ground-truth-related examples for better illustration.

\paragraph{Part I: Pairwise \textit{1v1} comparisons.}
We first compare \textbf{BiTDiff} with three representative baselines, each chosen to highlight a different failure mode.

\textbf{Ours vs.~Lodge~\cite{li2024lodge}.}
Compared with \textbf{BiTDiff}, \textbf{Lodge} often produces movements that are less consistent with realistic conducting gestures. Although the generated motion may remain plausible at a coarse body level, the gesture organization is often less convincing from a conducting perspective, especially in terms of beat articulation, upper-body coordination, and the overall visual logic of the conducting pattern. In contrast, \textbf{BiTDiff} produces motions that are more compatible with realistic conducting habits and better reflect the intended conducting structure.

\textbf{Ours vs.~DiffSHEG~\cite{chen2024diffsheg}.}
\textbf{DiffSHEG} tends to lose facial expressiveness and fine-grained hand motion details. This limitation is particularly noticeable in conducting, where hand articulation, wrist dynamics, and subtle facial cues are important for expressive communication. In the selected examples, \textbf{BiTDiff} preserves richer fine-grained details in both the hands and the face, resulting in motions that appear more expressive and visually complete.

\textbf{Ours vs.~DiffusionConductor~\cite{zhao2023taming}.}
\textbf{DiffusionConductor} frequently generates repetitive motion patterns. While such motions may remain temporally stable, they often lack sufficient variation and expressive development over time, making the conducting sequence appear mechanically repeated. By comparison, \textbf{BiTDiff} produces more diverse motion phrases and better captures the dynamic evolution of conducting behavior.

\paragraph{Part II: Multi-method \textit{1v1v1v1} comparisons.}
To provide a more comprehensive visual comparison, we additionally present two multi-method cases, each containing four methods under the same music segment. These examples allow readers to simultaneously compare differences in realism, expressiveness, fine-grained detail, and temporal diversity. Across these cases, \textbf{BiTDiff} generally shows more natural conducting patterns, richer hand and facial details, and less repetitive motion than the compared methods. Such side-by-side comparisons further support the quantitative results reported in the main paper.

\begin{figure*}[!t]
  \centering
  \includegraphics[width=0.75\linewidth]{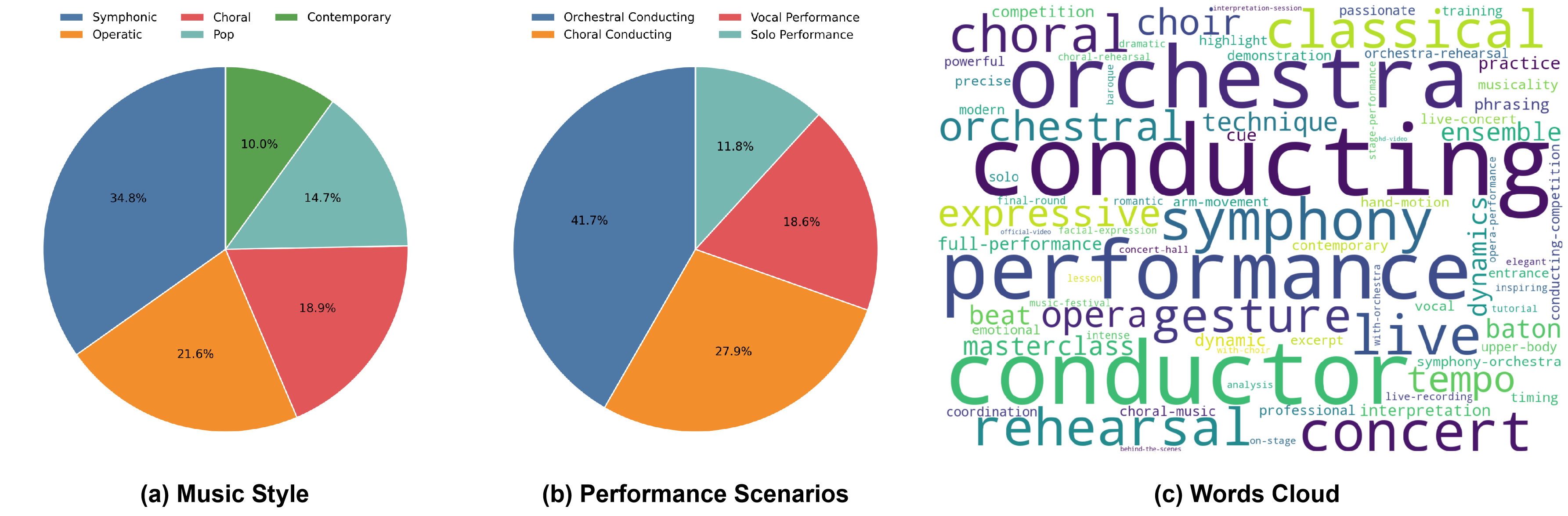}
  \vspace{-0.05in}
  \caption{\textbf{Dataset statistics of CM-Data.} (a) Distribution of musical styles. (b) Distribution of performance scenarios. (c) Word cloud of keywords collected from video titles and captions. These statistics illustrate the diversity of \textbf{CM-Data} in terms of musical content, performance setting, and textual metadata.}
  \label{fig:dataset}
  \vspace{-0.15in}
\end{figure*}

\subsection{Ablation}
We further provide qualitative ablation results to visualize the contribution of several important design choices in \textbf{BiTDiff}. Each example is about 10--15 seconds long. Similar to the comparison study, we primarily select representative cases from model predictions, so that the visual differences more directly reflect the effect of each architectural or objective-level modification.

\paragraph{\textit{w/o Vel} vs.~Ours.}
Removing the velocity-related constraint tends to make the generated motion appear less temporally smooth. In the selected examples, the resulting motion may exhibit slight jitter, especially in local transitions and small-amplitude movements. By contrast, the full model produces smoother and more temporally coherent conducting motion, indicating that the velocity-related design helps stabilize short-term dynamics.

\paragraph{\textit{Naive FK} vs.~Ours.}
Replacing our hand/body-decomposed kinematic constraint with a naive FK design mainly harms the quality of hand motion modeling. In the corresponding examples, hand motions may become weak, under-articulated, or even nearly static relative to the rest of the body. In contrast, \textbf{BiTDiff} better preserves fine-grained hand movement and produces more expressive conducting gestures, which qualitatively supports the necessity of the proposed decomposition strategy.

\paragraph{\textit{Mamba (uni)} vs.~Ours.}
Using unidirectional Mamba instead of BiMamba often leads to motion that is overly simple and less expressive. Although the generated sequence may remain broadly plausible, it tends to show weaker variation and reduced richness in temporal development. By comparison, the bidirectional design in \textbf{BiTDiff} helps produce conducting motions that are visually more dynamic and compositionally more complete.

\paragraph{\textit{Transformer} vs.~Ours.}
Compared with a Transformer-only temporal backbone, the qualitative difference is less dramatic than in some of the other ablations. In many cases, both variants can generate reasonable conducting motion. However, \textbf{BiTDiff} still tends to produce slightly better visual results, with somewhat more natural temporal transitions and finer local detail. More importantly, this variant should be interpreted together with the efficiency analysis in the main paper, where our BiMamba-Transformer design achieves substantially better generation speed while maintaining comparable or better motion quality.

Overall, these qualitative ablations provide intuitive evidence that the main components of \textbf{BiTDiff}---including the velocity-related design, the hand/body-decomposed FK constraint, and the BiMamba-based temporal modeling strategy---all contribute to the final generation quality.

\section{Dataset Details}
In total, \textbf{CM-Data} contains approximately 1{,}500 fine-grained 3D conducting motion samples, each with a duration ranging from 10 to 50 seconds, resulting in about 10 hours of SMPL-X motion data. To the best of our knowledge, \textbf{CM-Data} is the first and largest public dataset for 3D conducting motion generation. As shown in Fig.~\ref{fig:dataset}, \textbf{CM-Data} exhibits rich diversity in musical styles, performance scenarios, and title/caption keywords.

\textbf{CM-Data} exhibits several desirable properties for this task. We summarize its main characteristics as follows:

\begin{itemize}
    \item \textbf{Diverse performance scenarios.}
    \textbf{CM-Data} covers a broad range of music-related performance settings, including orchestral conducting, choral conducting, vocal performance, and solo performance scenarios. This diversity makes the dataset suitable for studying conducting motion under different musical and visual contexts.

    \item \textbf{Rich musical styles.}
    The dataset spans diverse musical styles, such as symphonic, operatic, choral, pop, and other contemporary music genres. As a result, it contains conducting motions associated with different rhythmic structures, expressive patterns, and performance conventions.

    \item \textbf{Diverse data sources.}
    \textbf{CM-Data} is curated from multiple online platforms, including YouTube, TikTok, and Douyin. This multi-source collection strategy increases the diversity of visual presentation, recording conditions, and performance contexts, which helps improve the representativeness of the dataset.

    \item \textbf{Variation across performers.}
    The dataset captures substantial variation in conducting style across different performers. This diversity is important because conducting is highly expressive in nature, and different conductors may exhibit noticeably different gesture habits even under similar musical content.

    \item \textbf{Fine-grained motion annotations.}
    Unlike coarse motion representations that mainly focus on global body movement, \textbf{CM-Data} provides fine-grained full-body, hand, and facial motion annotations. Such detailed annotations are essential for modeling the expressive and communicative nature of conducting.

    \item \textbf{A strong benchmark for future research.}
    Taken together, these characteristics establish \textbf{CM-Data} as a strong benchmark for fine-grained 3D conducting motion generation, supporting both high-quality motion synthesis and future research on expressive conducting analysis and controllable generation.
\end{itemize}

For evaluation, we randomly select 60 samples as the test set, while the remaining samples are used for training.

\section{User Study Questionnaire}

User evaluation is important for music-driven conducting motion generation because the perceived quality of conducting motion is inherently subjective and cannot be fully captured by automatic metrics alone~\cite{legrand2009perceiving,correia2024music}. Therefore, in addition to quantitative evaluation, we conduct a human study to assess the perceptual quality of the generated results from the perspective of evaluators with conducting-related backgrounds.

\subsection{Experimental Setup}

Following prior human-evaluation protocols in music-driven motion generation~\cite{yang2025megadance}, we randomly select 30 real-world music segments from the test distribution, each with a duration of 30 seconds, and generate corresponding conducting motion sequences using the compared methods. The evaluated methods include \textbf{BiTDiff} and the representative baselines reported in the main paper. For fair comparison, all methods are evaluated on the same set of music inputs.

For each music segment, the generated motion is rendered into a standardized video format for human assessment. To reduce irrelevant bias, all rendered videos use the same visual presentation settings, including the same character representation, camera viewpoint, playback speed, background style, and video resolution. The order of the videos is randomized for each participant, and method names are hidden throughout the study. Therefore, the evaluation is conducted in a double-blind manner: participants do not know which method produced a given sample, and the annotation interface does not reveal model identity during scoring.

The questionnaire is completed by 30 participants with conducting-related backgrounds. These participants are familiar with basic conducting principles such as beat patterns, gesture clarity, cueing, and expressive control, which makes them more suitable than general viewers for assessing the quality of conducting motion. Participants are compensated at a rate exceeding the local average hourly wage.

\subsection{Evaluation Criteria}

Each sample is evaluated on three dimensions using a 5-point Likert scale (\textit{Great}, \textit{Good}, \textit{Fair}, \textit{Bad}, \textit{Terrible}):

\begin{itemize}
    \item \textbf{Motion Synchronization (MS):} whether the conducting motion is well aligned with the rhythm, phrasing, and overall musical style of the input music.
    \item \textbf{Motion Fidelity (MF):} whether the generated motion is physically plausible, temporally stable, and consistent with natural human conducting behavior.
    \item \textbf{Motion Creativity (MC):} whether the motion is diverse, expressive, and compositionally rich, rather than repetitive or overly monotonous.
\end{itemize}

Before formal evaluation, participants are provided with concise written instructions explaining these three criteria, together with several illustrative examples to help them distinguish synchronization, physical plausibility, and creativity. During the annotation process, participants are asked to focus on the conducting motion itself rather than superficial rendering factors.

\subsection{Quality Control and Catch Trials}

To ensure annotation reliability, we include additional catch trials in the questionnaire. Specifically, the study contains both \textit{ground-truth} motion videos and intentionally \textit{distorted-motion} videos as quality-control samples. The ground-truth videos are expected to receive relatively high scores, while the distorted-motion videos are expected to receive relatively low scores. Participants who fail to consistently assign higher scores to the ground-truth videos and lower scores to the distorted-motion videos are excluded from the final analysis.

This design helps filter inattentive responses and improves the reliability of the collected human judgments. It is particularly important for conducting motion evaluation, where subtle differences in rhythmic control, articulation, and expressive quality may otherwise be difficult to assess reliably without careful viewing.

\subsection{Questionnaire Procedure}

In each trial, the participant watches a rendered conducting-motion video paired with its music audio and then rates the sample on the three criteria described above. Participants are allowed to replay the sample before submitting their ratings, so that they can make more reliable judgments on fine-grained details such as temporal stability, hand articulation, and music--motion correspondence.

To reduce order effects, fatigue effects, and method-specific presentation bias, the sample order is randomized. The full questionnaire is completed individually, and no participant is informed of the source method of any evaluated sample during the study.

For each method, we aggregate the participant ratings across all valid responses and report the mean scores for MS, MF, and MC in the main paper. These subjective results provide a complementary perspective to the automatic metrics and help evaluate whether the generated conducting motions are favorably perceived by human observers.

\subsection{Questionnaire Interface}
\begin{figure}[!t]
  \centering
  \includegraphics[width=0.75\linewidth]{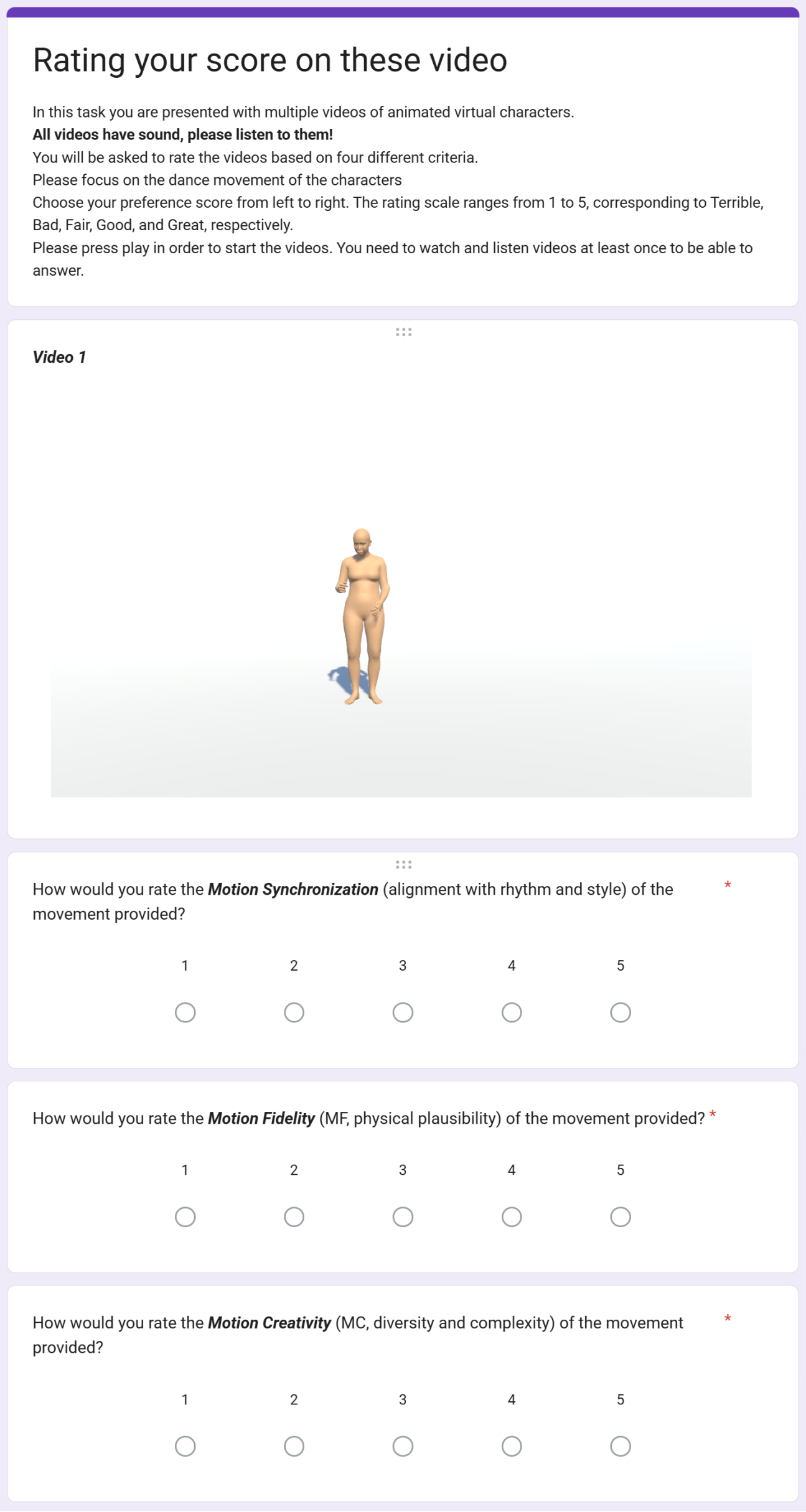}
  \vspace{-0.05in}
    \caption{Questionnaire interface used in our user study.}
  \label{fig:questionnaire}
  \vspace{-0.15in}
\end{figure}

Fig.~\ref{fig:questionnaire} shows the questionnaire interface used in our user study. For each trial, the interface presents the rendered video, the associated music, the three rating criteria, and the corresponding 5-point options. We include this figure to improve the transparency and reproducibility of our human-evaluation protocol.

\section{Limitation and Future Work}
\subsection{Customized 3D Motion Generation}
Although current music-driven conducting motion generation methods can produce plausible conducting sequences from music, they still provide limited support for personalized control. In real conducting practice, different conductors may interpret the same musical phrase with noticeably different gesture habits, including variations in beat shape, motion amplitude, cueing style, hand usage, upper-body engagement, and expressive timing. This variability is an important part of conducting as an artistic practice, but is not yet sufficiently modeled by existing systems.

Several future directions are worth exploring:
\begin{itemize}
    \item \textbf{Conductor-specific style modeling.} Future systems could explicitly model conductor-specific motion priors, so that the generated conducting motion better reflects individual gesture habits rather than only capturing a general conducting pattern.
    \item \textbf{Richer controllable conditions.} Beyond music input alone, it would be valuable to introduce additional controllable signals, such as textual instructions, partial motion guidance, or reference conducting clips, to support more flexible and user-steerable generation.
    \item \textbf{Personalized human-AI co-creation.} A more advanced direction is to develop interactive conducting-generation systems that can adapt to a user’s preferred conducting style and progressively refine the generated motion through human feedback.
\end{itemize}

This limitation is also related to a broader trend in the speech-driven gesture generation literature, where speaker-specific motion habits are considered important for realism and expressiveness. Similarly, exploring conductor-aware generation could be an important step toward more practical, expressive, and artistically meaningful human-AI conducting systems.

\subsection{Noise-Resistant 3D Motion Generation}
Another limitation lies in the quality of the reconstructed 3D conducting data. In particular, hand motions are frequently affected by self-occlusion, motion blur, and limited camera viewpoints, making accurate hand recovery from monocular videos inherently difficult. This issue is especially critical for conducting, because fine-grained hand articulation and continuous hand trajectories are central to musical communication.

From the data perspective, one major challenge is that monocular motion capture often struggles to recover subtle hand motion reliably. Even when the global body motion is reconstructed reasonably well, the hands may still contain noticeable noise, temporal instability, or missing details. This reconstruction difficulty directly affects the supervision quality of downstream generative models.

Possible future directions include:
\begin{itemize}
    \item \textbf{Stronger hand reconstruction pipelines.} Future work may develop more accurate and temporally stable hand-reconstruction methods specifically tailored to conducting videos, where the hands play a uniquely important role.
    \item \textbf{Sensor-assisted supervision.} A practical solution is to preserve a portion of sensor-captured motion data for model adaptation or fine-tuning, thereby providing higher-quality supervision for the most difficult motion components, especially the hands.
    \item \textbf{More robust generative modeling.} Another important direction is to design generation models that are more tolerant to noisy, incomplete, or partially inaccurate motion annotations during training, so that performance degrades more gracefully under imperfect supervision.
\end{itemize}

Improving robustness at both the data level and the model level would substantially enhance the reliability and generalization of future conducting motion generation systems.

\bibliographystyle{ACM-Reference-Format}
\bibliography{mm-appendix}